\documentclass[11pt,a4paper]{article}
\usepackage[hyperref]{emnlp-ijcnlp-2019}
\usepackage{times}
\usepackage{latexsym}
\usepackage{graphicx}
\usepackage{amsmath}
\usepackage{bm}
\usepackage{booktabs}

\usepackage{url}

\aclfinalcopy 

\title{Improving Fine-grained Entity Typing with Entity Linking}

\author{Hongliang Dai\textsuperscript{1}, Donghong Du\textsuperscript{1,3}, Xin Li\textsuperscript{2}, and Yangqiu Song\textsuperscript{1} \\
	\textsuperscript{1}Department of CSE, HKUST \\
	\textsuperscript{2}Tencent Technology (SZ) Co., Ltd. \\
	{\tt \textsuperscript{1}\{hdai,yqsong\}@cse.ust.hk } \\
    {\tt \textsuperscript{2}alonsoli@tencent.com}   \\
    {\tt \textsuperscript{3}dduaa@connect.ust.hk}\\}

\date{}

\begin{document}
\maketitle
\begin{abstract}
Fine-grained entity typing is a challenging problem since it usually involves a relatively large tag set and may require to understand the context of the entity mention.
In this paper, we use entity linking to help with the fine-grained entity type classification process. We propose a deep neural model that makes predictions based on both the context and the information obtained from entity linking results. Experimental results on two commonly used datasets demonstrates the effectiveness of our approach. On both datasets, it achieves more than 5\% absolute strict accuracy improvement over the state of the art.
\end{abstract}

\section{Introduction}

Given a piece of text and the span of an entity mention in this text, fine-grained entity typing (FET) is the task of assigning fine-grained type labels to the mention \cite{ling2012fine}. The assigned labels should be context dependent \cite{gillick2014context}. For example, in the sentence ``Trump threatens to pull US out of World Trade Organization,'' the mention ``Trump'' should be labeled as /person and /person/politician, although Donald Trump also had other occupations such as businessman, TV personality, etc.

This task is challenging because it usually uses a relatively large tag set, and some mentions may require the understanding of the context to be correctly labeled. Moreover, since manual annotation is very labor-intensive, existing approaches have to rely on distant supervision to train models \cite{ling2012fine,ghaddar2018transforming}.

Thus, the use of extra information to help with the classification process becomes very important. In this paper, we improve FET with entity linking (EL). EL is helpful for a model to make typing decisions because if a mention is correctly linked to its target entity, we can directly obtain the type information about this entity in the knowledge base (KB).
For example, in the sentence ``There were some great discussions on a variety of issues facing Federal Way,'' the mention ``Federal Way'' may be incorrectly labeled as a company by some FET models. Such a mistake can be avoided after linking it to the city Federal Way, Washington.
For cases that require the understanding of the context, using entity linking results is also beneficial. In the aforementioned example where ``Trump'' is the mention, obtaining all the types of Donald Trump in the knowledge base (e.g., politician, businessman, TV personality, etc.) is still informative for inferring the correct type (i.e., politician) that fits the context, since they narrows the possible labels down.

However, the information obtained through EL should not be fully trusted since it is not always accurate. Even when a mention is correctly linked to an entity, the type information of this entity in the KB may be incomplete or outdated. Thus, in this paper, we propose a deep neural fine-grained entity typing model that flexibly predicts labels based on the context, the mention string, and the type information from KB obtained with EL.

Using EL also introduces a new problem for the training process. Currently, a widely used approach to create FET training samples is to use the anchor links in Wikipedia \cite{ling2012fine,ren2016afet}. Each anchor link is regarded as a mention, and is weakly labeled with all the types of its referred entity (the Wikipedia page the anchor link points to) in KB. Our approach, when links the mention correctly, also uses all the types of the referred entity in KB as extra information. This may cause the trained model to overfit the weakly labeled data. We design a variant of the hinge loss and introduce noise during training to address this problem.


We conduct experiments on two commonly used FET datasets. Experimental results show that introducing information obtained through entity linking and having a deep neural model both helps to improve FET performance. Our model achieves more than 5\% absolute strict accuracy improvement over the state of the art on both datasets. 

Our contributions are summarized as follows: 

\begin{itemize}
	\item We propose a deep neural fine-grained entity typing model that utilizes type information from KB obtained through entity linking.
	\item We address the problem that our model may overfit the weakly labeled data by using a variant of the hinge-loss and introducing noise during training.
	\item We demonstrate the effectiveness of our approach with experimental results on commonly used FET datasets.
\end{itemize}

Our code is available at \url{https://github.com/HKUST-KnowComp/IFETEL}.

\section{Related Work}

An early effort of classifying named entities into fine-grained types can be found in \cite{fleischman2002fine}, which only focuses on person names. Latter, datasets with larger type sets are constructed \cite{weischedel2005,ling2012fine,choi2018ultra}. These datasets are more preferred by recent studies \cite{ren2016afet,murty2018hierarchical}.

Most of the existing approaches proposed for FET are learning based. The features used by these approaches can either be hand-crafted \cite{ling2012fine,gillick2014context} or learned from neural network models \cite{shimaoka2017neural,xu2018neural,xin2018improving}. Since FET systems usually use distant supervision for training, the labels of the training samples can be noisy, erroneous or overly specific. Several studies \cite{ren2016label,xin2018put,xu2018neural} address these problems by separating clean mentions and noisy mentions, modeling type correction \cite{ren2016afet}, using a hierarchy-aware loss \cite{xu2018neural}, etc.

\cite{huang2016building} and \cite{zhou2018zero} are two studies that are most related to this paper. \citet{huang2016building} propose an unsupervised FET system where EL is an importat component. But they use EL to help with clustering and type name selection, which is very different from how we use it to improve the performance of a supervised FET model. \cite{zhou2018zero} finds related entities based on the context instead of directly applying EL. The types of these entities are then used for inferring the type of the mention.


\begin{figure*}
\centering
	\includegraphics[width=0.99\textwidth]{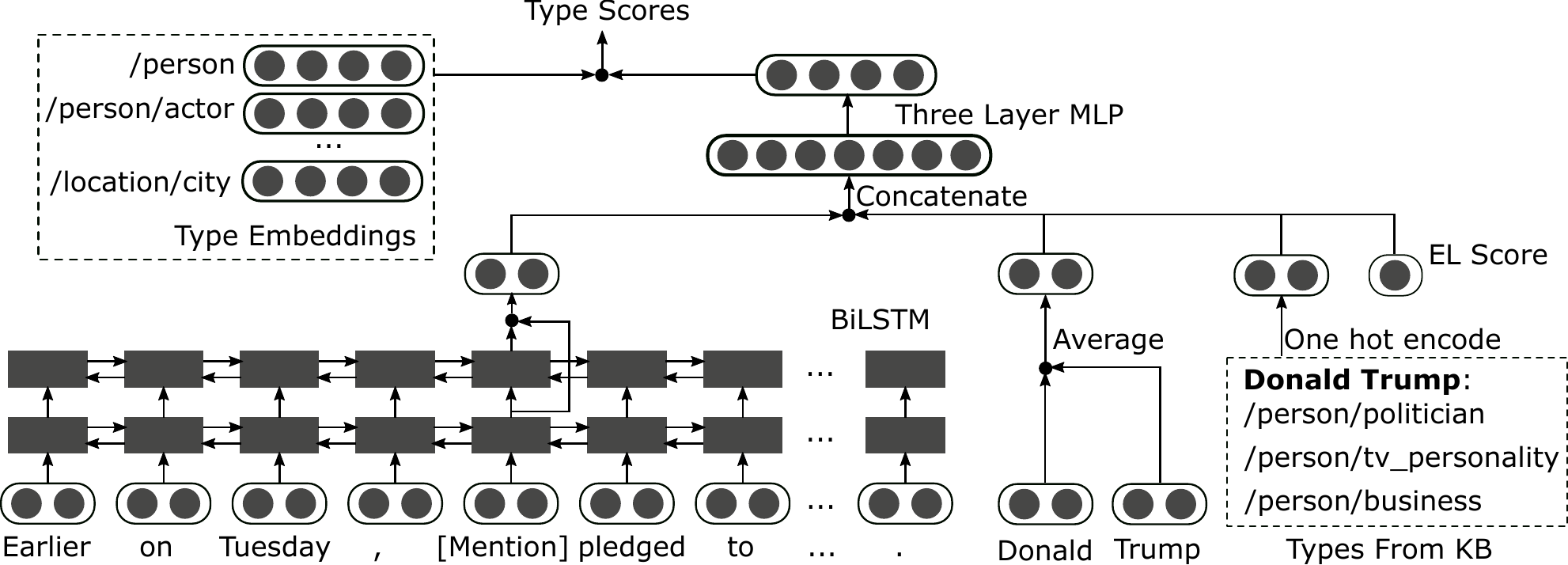}
\caption{Our approach. The example sentence is ``Earlier on Tuesday, \textit{Donald Trump} pledged to help hard-hit U.S. farmers caught in the middle of the escalating trade war.'' Here, the correct label for the mention \textit{Donald Trump} should be /person, /person/politician. ``[Mention]'' is a special token that we use to represent the mention.}
	\label{fig:method}
\end{figure*}

\section{Method}

Let $T$ be a predefined tag set, which includes all the types we want to assign to mentions. Given a mention $m$ and its context, the task is to predict a set of types $\bm{\tau}\subset T$ suitable for this mention. Thus, this is a multi-class, multi-label classification problem \cite{ling2012fine}. Next, we will introduce our approach for this problem in detail, including the neural model, the training of the model, and the entity linking algorithm we use.


\subsection{Fine-grained Entity Typing Model}
\label{sec:fet-model}

\paragraph{Input} Each input sample to our FET system contains one mention and the sentence it belongs to. We denote $w_1,w_2,...,w_n$ as the words in the current sentence, $w_{p_1},w_{p_2},...,w_{p_l}$ as the words in the mention string, where $n$ is the number of words in the sentence, $p_1,...,p_l$ are the indices of the words in the mention string, $l$ is the number of words in the mention string. We also use a set of pretrained word embeddings.

Our FET approach is illustrated in Figure \ref{fig:method}. It first constructs three representations: \textit{context representation}, \textit{mention string representation}, and \textit{KB type representation}. Note that the KB type representation is obtained from a knowledge base through entity linking and is independent of the context of the mention.

\paragraph{Context Representation} To obtain the context representation, we first use a special token $w_m$ to represent the mention (the token ``[Mention]'' in Figure \ref{fig:method}). Then, the word sequence of the sentence becomes $w_1,...,w_{p_l-1},\allowbreak w_m,\allowbreak w_{p_l+1},...,w_n$. Their corresponding word embeddings are fed into two layers of BiLSTMs. Let $\bm{h}_m^1$ and $\bm{h}_m^2$ be the output of the first and the second layer of BiLSTMs for $w_m$, respectively. We use $\bm{f}_c=\bm{h}_m^1+\bm{h}_m^2$ as the context representation vector.

\paragraph{Mention String Representation} Let $\bm{x}_1,...,\bm{x}_l$ be the word embeddings of the mention string words $w_{p_1},...,w_{p_l}$. Then the mention string representation $\bm{f}_s=(\sum_{i=1}^l \bm{x}_i)/l$.

\paragraph{KB Type Representation} To obtain the KB type representation, we run an EL algorithm for the current mention. If the EL algorithm returns an entity, we retrieve the types of of this entity from the KB. We use Freebase as our KB\footnote{We use Freebase mainly because it is widely used by existing studies. Wikidata is an alternative.}. Since the types in Freebase is different from $T$, the target type set, they are mapped to the types in $T$ with rules similar to those used in \cite{zhou2018zero}. Afterwards, we perform one hot encoding on these types to get the KB Type Representation $\bm{f}_e$. If the EL algorithm returns NIL (i.e., the mention cannot be linked to an entity), we simply one hot encode the empty type set.

\paragraph{Prediction} Apart from the three representations, we also obtain the score returned by our entity linking algorithm, which indicates its confidence on the linking result. We denote it as a one dimensional vector $\bm{g}$. Then, we get $\bm{f}=\bm{f}_c\oplus\bm{f}_s\oplus \bm{f}_e\oplus \bm{g}$, where $\oplus$ means concatenation. $\bm{f}$ is then fed into an MLP that contains three dense layers to obtain $\bm{u}_m$, out final representation for the current mention sample $m$. Let $t_1,t_2,...,t_k$ be all the types in $T$, where $k=|T|$. We embed them into the same space as $\bm{u}_m$ by assigning each of them a dense vector \cite{yogatama2015embedding}. These vectors are denoted as $\bm{t}_1,...,\bm{t}_k$. Then the score of the mention $m$ having the type $t_i\in T$ is calculated as the dot product of $\bm{u}_m$ and $\bm{t}_i$:
\begin{equation}
    s(m,t_i)=\bm{u}_m \cdot \bm{t}_i.
\end{equation}
We predict $t_i$ as a type of $m$ if $s(m,t_i)>0$.

\subsection{Model Training}

Following existing studies, we also generate training data by using the \textit{anchor links} in Wikipedia. Each anchor link can be used as a mention. These mentions are labeled by mapping the Freebase types of the target entries to the tag set $T$ \cite{ling2012fine}.

Since the \textit{KB type representations} we use in our FET model are also obtained through mapping Freebase types, they will perfectly match the automatically generated labels for the mentions that are correctly linked (i.e., when the entity returned by the EL algorithm and the target entry of the anchor link are the same). For example, in Figure \ref{fig:method}, suppose the example sentence is a training sample obtained from Wikipedia, where ``Donald Trump'' is an anchor link points to the Wikipedia page of \textit{Donald Trump}. After mapping the Freebase types of \textit{Donald Trump} to the target tag set, this sample will be weakly annotated as /person/politician, /person/tv\_personality, and /person/business, which is exactly the same as the type information (the ``Types From KB'' in Figure \ref{fig:method}) obtained through EL. Thus, during training, when the EL system links the mention to the correct entity, the model only needs to output the types in the \textit{KB type representation}. This may cause the trained model to overfit the weakly labeled training data. For most types of entities such as locations and organizations, it is fine since they usually have the same types in different contexts. But it is problematic for person mentions, as their types can be context dependent.

To address this problem, during training, if a mention is linked to a person entity by our entity linking algorithm, we add a random fine-grained person type label that does not belong to this entity while generating the \textit{KB type representation}. For example, if the mention is linked to a person with types /person/actor and /person/author, a random label /person/politician may be added. This will force the model to still infer the type labels from the context even when the mention is correctly linked, since the \textit{KB type representation} no longer perfectly match the weak labels. 

To make it more flexible, we also propose to use a variant of the hinge loss used by \cite{abhishek2017fine} to train our model:
\begin{equation}
\begin{split}
    L&=\sum_m [\sum_{t\in \tau_m}\max(0,1-s(m,t)) \\
     &+\sum_{t\in \bar{\tau}_m} \lambda (t) \max(0,1+s(m,t))]
\end{split}
\end{equation}
where $\tau_m$ is the correct type set for mention $m$, $\bar{\tau}_m$ is the incorrect type set. $\lambda (t)\in [1,+\infty)$ is a predefined parameter to impose a larger penalty if the type $t$ is incorrectly predicted as positive. Since the problem of overfitting the weakly annotated labels is more severe for person mentions, we set $\lambda (t)=\lambda_P$ if $t$ is a fine-grained person type, and $\lambda (t)=1$ for all other types.

During training, we also randomly set the EL results of half of the training samples to be NIL. So that the model can perform well for mentions that cannot be linked to the KB at test time.

\subsection{Entity Linking Algorithm}
In this paper, we use a simple EL algorithm that directly links the mention to the entity with the greatest commonness score. Commonness \cite{pan2015unsupervised,medelyan2008integrating} is calculated base on the anchor links in Wikipedia. It estimates the probability of an entity given only the mention string. In our FET approach, the commonness score is also used as the confidence on the linking result (i.e., the $\bm{g}$ used in the prediction part of Subsection \ref{sec:fet-model}). Within a same document, we also use the same heuristic used in \cite{ganea2017deep} to find coreferences of generic mentions of persons (e.g., ``Matt'') to more specific mentions (e.g., ``Matt Damon''). 

We also tried other more advanced EL methods in our experiments. However, they do not improve the final performance of our model. Experimental results of using the EL system proposed in \cite{ganea2017deep} is provided in Section \ref{sec:exp}.

\section{Experiments}
\label{sec:exp}

\begin{table*}[t!]
\begin{center}
\begin{tabular}{l|c|c|c|c|c|c}
\toprule
Dataset & \multicolumn{3}{c|}{FIGER (GOLD)} & \multicolumn{3}{c}{BBN} \\ \hline
Approach & Accuracy & Macro F1 & Micro F1 & Accuracy & Macro F1 & Micro F1 \\ \midrule
AFET & 53.3 & 69.3 & 66.4 & 67.0 & 72.7 & 73.5 \\
AAA & 65.8 & 81.2 & 77.4 & 73.3 & 79.1 & 79.2 \\
NFETC & 68.9 & 81.9 & 79.0 & 72.1 & 77.1 & 77.5 \\
CLSC & - & - & - & 74.7 & 80.7 & 80.5 \\ \hline
Ours (NonDeep NoEL) & 65.9 & 81.7 & 78.0 & 69.3 & 81.4 & 81.5 \\
Ours (NonDeep) & 72.3 & 85.4 & 82.6 & 79.1 & 87.9 & 88.4 \\
Ours (DirectTrain) & 69.1 & 85.2 & 82.2 & - & - & - \\
Ours (NoEL) & 69.8 & 82.7 & 80.4 & 80.5 & 87.5 & 88.0 \\
Ours (LocAttEL) & 75.1 & 86.3 & 83.9 & \textbf{82.8} & 88.9 & 89.5 \\
Ours (Full) & \textbf{75.5} & \textbf{87.1} & \textbf{84.6}  & 82.5 & \textbf{89.2} & \textbf{89.6} \\
\bottomrule
\end{tabular}
\end{center}
\caption{\label{tab:fet-perf} Fine-grained entity typing performance. The performance of ``Ours (DirectTrain)'' on BBN is omitted since this dataset does not have fine-grained types for person.}
\end{table*}

\subsection{Setup}
We use two datasets: FIGER (GOLD) \cite{ling2012fine} and BBN \cite{weischedel2005}. The sizes of their tag sets are 113 and 47, respectively. FIGER (GOLD) allows mentions to have multiple type paths, but BBN does not. Another commonly used dataset, OntoNotes \cite{gillick2014context}, is not used since it contains many pronoun and common noun phrase mentions such as ``it,'' ``he,'' ``a thrift institution,'' which are not suitable to directly apply entity linking on.

Following \cite{ling2012fine}, we generate weakly labeled datasets for training with Wikipedia anchor links. Since the tag sets used by FIGER (GOLD) and BBN are different, we create a training set for each of them. For each dataset, $2,000$ weakly labeled samples are randomly picked to form a development set. We also manually annotated 50 person mentions collected from news articles for tuning the parameter $\lambda_P$.

We use the 300 dimensional pretrained GloVe word vectors provided by \cite{pennington2014glove}. The hidden layer sizes of the two layers of BiLSTMs are both set to 250. For the three-layer MLP, the size of the two hidden layers are both set to 500. The size of the type embeddings is 500. $\lambda_P$ is set to 2.0. We also apply batch normalization and dropout to the input of each dense layer in our three-layer MLP during training.

We use strict accuracy, Macro F1, and Micro F1 to evaluate fine-grained typing performance \cite{ling2012fine}.

\subsection{Compared Methods}
We compare with the following existing approaches: AFET \cite{ren2016afet}, AAA \cite{abhishek2017fine}, NFETC \cite{xu2018neural}, and CLSC \cite{chen2019improving}.

We use \textbf{Ours (Full)} to represent our full model, and also compare with five variants of our own approach: \textbf{Ours (DirectTrain)} is trained without adding random person types while obtaining the KB type representation, and $\lambda_P$ is set to 1; \textbf{Ours (NoEL)} does not use entity linking, i.e., the KB type representation and the entity linking confidence score are removed, and the model is trained in DirectTrain style; \textbf{Ours (NonDeep)} uses one BiLSTM layer and replaces the MLP with a dense layer; \textbf{Ours (NonDeep NoEL)} is the NoEL version of \textit{Ours (NonDeep)}; \textbf{Ours (LocAttEL)} uses the entity linking approach proposed in \cite{ganea2017deep} instead of our own commonness based approach. \textit{Ours (Full)}, \textit{Ours (DirectTrain)}, and \textit{Ours (NonDeep)} all use our own commonness based entity linking approach.

\subsection{Results}
The experimental results are listed in Table \ref{tab:fet-perf}. As we can see, our approach performs much better than existing approaches on both datasets. 

The benefit of using entity linking in our approach can be verified by comparing \textit{Ours (Full)} and \textit{Ours (NoEL)}. The performance on both datasets decreases if the entity linking part is removed. Especially on FIGER (GOLD), the strict accuracy drops from 75.5 to 69.8. Using entity linking improves less on BBN. We think this is because of three reasons: 1) BBN has a much smaller tag set than FIGER (GOLD); 2) BBN does not allow a mention to be annotated with multiple type paths (e.g., labeling a mention with both /building and /location is not allowed), thus the task is easier; 3) By making the model deep, the performance on BBN is already improved a lot, which makes further improvement harder. 

The improvement of our full approach over \textit{Ours (DirectTrain)} on FIGER (GOLD) indicates that the techniques we use to avoid overfitting the weakly labeled data are also effective.

\textit{Ours (LocAttEL)}, which uses a more advanced EL system, does not achieve better performance than \textit{Ours (Full)}, which uses our own EL approach. After manually checking the results of the two EL approaches and the predictions of our model on FIGER (GOLD), we think this is mainly because: 1) Our model also uses the context while making predictions. Sometimes, if it ``thinks'' that the type information provided by EL is incorrect, it may not use it. 2) The performances of different EL approaches also depends on the dataset and the types of entities used for evaluation. We find that on FIGER (GOLD), the approach in \cite{ganea2017deep} is better at distinguishing locations and sports teams, but it may also make some mistakes that our simple EL method does not. For example, it may incorrectly link ``March,'' the month, to an entity whose Wikipedia description fits the context better. 3) For some mentions, although the EL system links it to an incorrect entity, the type of this entity is the same with the correct entity.

\section{Conclusions}
We propose a deep neural model to improve fine-grained entity typing with entity linking. The problem of overfitting the weakly labeled training data is addressed by using a variant of the hinge loss and introducing noise during training. We conduct experiments on two commonly used dataset. The experimental results demonstrates the effectiveness of our approach.

\section*{Acknowledgments}
This paper was supported by the Early Career Scheme (ECS, No. 26206717) from Research Grants Council in Hong Kong and WeChat-HKUST WHAT Lab on Artificial Intelligence Technology.

\bibliography{emnlp-ijcnlp-2019}
\bibliographystyle{acl_natbib}

\end{document}